\documentclass[letterpaper, 10 pt, conference]{ieeeconf}  

\usepackage[hidelinks]{hyperref}
\usepackage{amsmath}
\usepackage{graphicx}
\usepackage{stfloats}
\usepackage{float}
\usepackage{booktabs}
\usepackage{siunitx}
\usepackage{amssymb}
\usepackage{multirow}
\usepackage{pgfplots}
\usepackage{cleveref}
\usepackage{subcaption}  
\pgfplotsset{compat=1.18}
\usepackage{algorithm}
\usepackage[noend]{algpseudocode}
\usepackage{tikz}
\usetikzlibrary{calc} 
\usetikzlibrary{shapes.geometric, arrows.meta, positioning,matrix}
\usetikzlibrary{positioning, arrows.meta}
\tikzset{
  node/.style={circle, draw=black, thick, minimum size=1.5cm, align=center},
  graynode/.style={circle, draw=gray, fill=gray!10, thick, minimum size=1.5cm, align=center},
  arrow/.style={-{Stealth}, thick},
}
\usepackage{makecell} 

\usepackage[style=ieee, maxbibnames=99, citestyle=numeric-comp, doi=false,url=false, natbib=true,natbib=true]{biblatex}
\AtEveryBibitem{\clearfield{issn}}
\AtEveryCitekey{\clearfield{issn}}
\AtEveryBibitem{%
  \clearfield{note}%
}

\IEEEoverridecommandlockouts                              

\usepackage{graphics} 
\usepackage{hyperref}
\usepackage{amsmath} 
\title{\LARGE \bf
Adaptive Science Operations in Deep Space Missions \\ Using Offline Belief State Planning}

\author{
Grace Ra Kim\thanks{$^{1}$ Department of Aeronautics and Astronautics, Stanford University, Stanford, CA, 94305, USA  
\href{mailto:gkim65@stanford.edu}{\tt\small gkim65@stanford.edu}
\href{mailto:hlwarner@stanford.edu}{\tt\small hlwarner@stanford.edu}
}\textsuperscript{1}\textsuperscript{*},  Hailey Warner\textsuperscript{1}\textsuperscript{*},
Duncan Eddy\textsuperscript{1},
Evan Astle\thanks{$^{2}$ Intelligent Systems Division, NASA Ames Research Center, Moffett Field, CA, 94035, USA 
}\textsuperscript{2},\\ 
Zachary Booth\textsuperscript{2}, 
Edward Balaban\textsuperscript{2}, Mykel J. Kochenderfer\textsuperscript{1}\\
\small{\textsuperscript{*}Equal contribution.}}

\begin{document}

\maketitle
\thispagestyle{empty}
\pagestyle{empty}

\begin{abstract}

Deep space missions face extreme communication delays and environmental uncertainty that prevent real-time ground operations. 
To support autonomous science operations in communication-constrained environments, we present a partially observable Markov decision process (POMDP) framework that adaptively sequences spacecraft science instruments.
We integrate a Bayesian network into the POMDP observation space to manage the high-dimensional and uncertain measurements typical of astrobiology missions. This network compactly encodes dependencies among measurements and improves the interpretability and computational tractability of science data. Instrument operation policies are computed offline, allowing resource-aware plans to be generated and thoroughly validated prior to launch. We use the Enceladus Orbilander's proposed Life Detection Suite (LDS) as a case study, demonstrating how Bayesian network structure and reward shaping influence system performance. We compare our method against the mission’s baseline Concept of Operations (ConOps), evaluating both misclassification rates and performance in off-nominal sample accumulation scenarios. Our approach reduces sample identification errors by nearly 40\%. 


\end{abstract}

\section{Introduction}

Deep space exploration in extreme environments requires autonomous systems capable of maximizing science return under harsh conditions without human-in-the-loop operations. Missions such as NASA's Europa Clipper~\cite{howell_nasas_2020} and ESA's JUICE~\cite{fletcher_jupiter_2023} face Earth-communication delays exceeding one hour~\cite{bushnell2021futures}, preventing timely human intervention. In astrobiological missions, biosignatures such as cell membranes can degrade within hours to days of sampling~\cite{mathies2017feasibility},~\cite{boulesteix2025icar}. Testing delays risk the irretrievable loss of scientific data, potentially compromising mission objectives. This challenge stresses the need for autonomous systems capable of making timely, critical decisions onboard to maximize science return.

Spacecraft autonomy has advanced across multiple domains, enhancing real-time data processing, navigation, trajectory planning, and fault detection~\cite{banerjee2023resiliency}. For example, the Perseverance rover demonstrated autonomous navigation and target selection on Mars~\cite{mars_perserverance}, and the Deep-space Autonomous Robotic Explorer (DARE) optimized trajectories for asteroid reconnaissance~\cite{echigo_autonomy_2024}. 
Despite these advancements, most existing autonomy architectures remain limited to foundational operational functions such as mobility and resource management rather than scientific decision-making. To enable the adoption of advanced autonomy in science operations, gaining the trust of scientific stakeholders is essential through efficient, transparent, and verifiable techniques that reflect expert preferences. Offline, precomputed decision policies offer a pathway to building such autonomous systems, as they allow mission operators to rigorously validate behavior prior to deployment and ensure alignment with mission objectives and constraints.

To address this need, we build upon a partially observable Markov decision process (POMDP), a mathematical framework for sequential decision-making under uncertainty and partial observability. This framework has been successfully used in various space applications, including collision avoidance~\cite{bourriez2023spacecraft} and planetary navigation~\cite{balaban2018system}. POMDPs use probabilistic observation models to maintain beliefs about uncertain environments and optimize long-term objectives.

In this work, we use a POMDP framework to automate operation of the Enceladus Orbilander life detection instruments. We structure the observation model as a Bayesian network to efficiently capture complex correlations between biosignatures. An offline POMDP solver, SARSOP~\cite{kurniawati2008sarsop}, is used to precompute optimal instrument-use policies in the form of a lookup table. These policies can be thoroughly validated prior to launch while still adapting in real time to instrument findings and off-nominal events. To evaluate our method, we compare its performance against the baseline Orbilander concept of operations using metrics such as life detection false positive rates, false negative rates, and performance in off-nominal scenarios. Our results demonstrate up to a 40\% reduction in identification errors compared to the baseline, underscoring the benefits of offline belief state planning.
\section{Enceladus Orbilander Mission Concept}


We seek to automate operation of the Life Detection Suite (LDS) of the Enceladus Orbilander, a flagship mission concept from the 2023–2032 Planetary Science Decadal Survey~\cite{decadal}. Our goal is to develop a decision-making framework that autonomously sequences LDS instruments to maximize science return while satisfying mission constraints.

The Enceladus Orbilander, developed by NASA and the Johns Hopkins Applied Physics Laboratory (APL), is designed to assess the habitability of Enceladus and search for evidence of life. This Saturnian moon has a subsurface ocean that ejects ice material through surface plumes. Many researchers speculate these plumes contain bioactive material, making Enceladus a prime target for astrobiological exploration~\cite{mathies2017feasibility},~\cite{new_enceladus_sampling_2021}. The Orbilander mission architecture combines an orbiter and a lander, passively collecting samples in a funnel during plume fly-bys and actively sampling the surface using a robotic arm scoop. To detect biosignatures, the spacecraft carries a sophisticated suite of six instruments: two mass spectrometers, an electrochemical sensor array, an organic analyzer, a microscope, and a nanopore sequencer. These instruments are capable of detecting a wide range of potential biosignatures\textemdash from amino acids and cells to weak indicators like pH (\Cref{tab:biosignatures}). During orbit, the spacecraft is expected to collect between 1 and 20 samples depending on plume activity~\cite{mackenzie2020enceladus}.

Several mission constraints complicate LDS operations. The spacecraft’s 12-hour orbital period limits the duration and frequency of Earth communication windows, reducing opportunities for real-time human decision-making. Power is also limited: the spacecraft's radioisotope thermoelectric generators (RTGs) are projected to degrade by 13\% before Saturn orbit insertion and up to 22\% by landing~\cite{mackenzie2020enceladus}. 

These limitations drive the need for robust and timely decision-making without reliance on ground support. Offline autonomous solutions must adapt not only to environmental uncertainty, but also constrained time and resources.



\section{Methodology}

This section describes the general formulation of a POMDP~\cite{spaan2012partially} and how we model the Life Detection Suite of the Enceladus Orbilander within this framework. To efficiently structure sensor observations, we incorporate a Bayesian network into the observation model. We then use SARSOP~\cite{kurniawati2008sarsop}, an approximate offline POMDP solver, to compute policies that map the vehicle's current state and observations to the next instrument action. These policies are optimized to maximize biosignature detection accuracy while maintaining robustness under off-nominal conditions.

A partially observable Markov decision process is a mathematical framework for modeling sequential decision-making problems under uncertainty~\cite{kochenderfer2022algorithms}. A POMDP $\mathcal{P}$ is 
defined as the tuple 
\begin{equation}
    \mathcal{P} = (\mathcal{S}, \mathcal{A}, \mathcal{O},T, O, R)
\end{equation}
containing the model's state, action, and observation spaces along with transition, observation, and reward functions. In a POMDP, an agent (e.g. the spacecraft) interacts with an environment (e.g. a planetary surface) characterized by a set of hidden states $\mathcal{S}$. The agent receives a noisy observation $o \in \mathcal{O}$ of the environment with likelihood $O(o \mid s, a)$. Based on $o$, the agent selects an action $a \in \mathcal{A}$ to perform. The state then transitions to $s' \in \mathcal{S}$ with likelihood $T(s' \mid s, a)$. After transitioning, the agent receives a reward $R(s, a)$, which quantifies the immediate benefit of that state and action. 



We next formulate the Enceladus Orbilander instrument operations as a POMDP. By narrowing our problem's scope to LDS operations, we reduce the complexity of the state, action, and observation spaces without affecting other subsystems.

\subsection{States} We define the POMDP state vector as $s = [s_V, s_L]$. Here, $s_V$ denotes the volume of sample available for testing in the Orbilander's funnel. We assume this value is fully observable, meaning that observations $o_V=s_V$ with no sensor error. The second state variable $s_L$ represents the sample's true life state\textemdash that is, whether the sample inside the LDS preparation chamber contains any biotic material. State $s_L$ is only partially observable via instrument measurements, which produce noisy observations $o_L$ of the characteristics outlined in \Cref{tab:biosignatures}.

\subsection{Observations} POMDPs can grow quickly in complexity due to high-dimensional observation spaces, which can make solving them computationally intractable. To address this issue, we use a Bayesian network to compactly model observations. A Bayesian network is a probabilistic model that represents a set of random variables and their conditional dependencies using a directed, acyclic graph. This structure is suited to astrobiological settings, where multiple sensors may find correlated observations of the same underlying processes. By factoring the joint distribution into a product of conditional probability densities, Bayesian networks drastically reduce the number of parameters required to model samples. For example, a Bayesian network with structure $A \rightarrow B \rightarrow C$ induces a joint probability distribution that can be factored as $P(A, B, C) = P(A)\cdot P(B\mid A) \cdot P(C\mid B)$. For an overview of Bayesian networks, see~\citet{koller2009pgm} or~\citet{kochenderfer2022algorithms}.

The structure of Bayesian networks can be learned from data or designed by experts. We adopt an expert-driven approach due to limited data and epistemic uncertainty regarding biotic characteristics in deep space.

To build a Bayesian network that represents the relevant biosignatures and habitability characteristics on Enceladus, we select random variables that match measurement capabilities of the Orbilander instrument suite. \Cref{tab:biosignatures} outlines a full list of measurements and biosignatures expected from Enceladus plume samples. Generally, a strong biosignature is one where the likelihood of a certain measurement is far greater when biotic processes are present than when only abiotic processes are involved; that is, $ P(o\mid\text{biotic})\gg P(o\mid\text{abiotic})$. Additionally, biotic signals must be strong enough to exceed the detection threshold of each instrument. 

\begin{table}[t]
  \centering
  \caption{Selected Biosignatures and Habitability Metrics}
  \label{tab:biosignatures}
  \begin{tabular}{@{} l l c c @{}}
    \toprule
    Name & Characteristic                           & Values                         & Parent Nodes \\ 
    \midrule
    $s_L$  & Life                                   & \{0,1\}                       & - \\
    $o_1$  & Polyelectrolyte Presence               & \{0,1\}                       & $s_L$ \\
    $o_2$  & Cell Membrane Presence                 & \{0,1\}                       & $s_L$ \\
    $o_3$  & Autofluorescence                       & \{0,1\}                       & $s_L$ \\
    $o_4$  & Molecular Assembly Index $\geq 15$     & \{0,1\}                       & $s_L$ \\
    $o_5$  & Biotic Amino Acid Diversity            & \{0, \ldots, 22\} & $s_L$ \\
    $o_6$  & L:R Chirality Ratio (\%)                          & [0, 100]                    & $s_L$ \\
    $o_7$  & Salinity (\%)                              & [0, ..., 100]                   & $o_2$ \\
    $o_8$ & CHNOPS Abundance (\%)                       & [0, ..., 100]                    & $o_4, o_5$ \\
    $o_9$  & pH                                     & $[0, 14]$ & $o_1, o_5$ \\
    $o_{10}$  & Redox Potential [V]                 & $ [-0.5, 0]$ & $o_5$ \\
    \bottomrule
  \end{tabular}
\end{table}

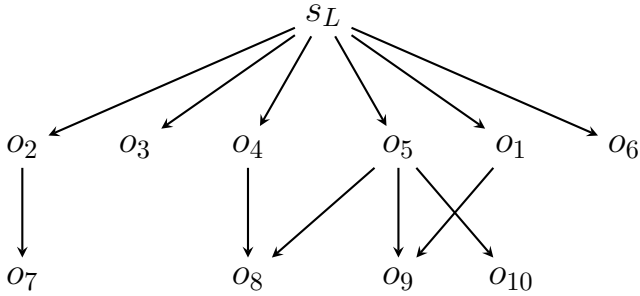
\begin{figure}[tbp!]
    \centering
    \begin{tikzpicture}[
  ->, >=stealth, thick,
  every node/.style={font=\Large}
]
\node (C0) at (0,0) {$s_L$};

\node (C2) at (-4,-1.75) {$o_2$};
\node (C3) at (-2.5,-1.75) {$o_3$};
\node (C4) at (-1,-1.75) {$o_4$};
\node (C5) at (1,-1.75) {$o_5$};
\node (C1) at (2.5,-1.75) {$o_1$};
\node (C6) at (4,-1.75) {$o_6$};

\node (C7) at (-4,-3.5) {$o_7$};
\node (C8) at (-1,-3.5) {$o_8$};
\node (C9) at (1,-3.5) {$o_9$};
\node (C10) at (2.5,-3.5) {$o_{10}$};

\draw (C0) -- (C2);
\draw (C0) -- (C3);
\draw (C0) -- (C4);
\draw (C0) -- (C5);
\draw (C0) -- (C1);
\draw (C0) -- (C6);

\draw (C2) -- (C7);
\draw (C4) -- (C8);
\draw (C5) -- (C8);
\draw (C5) -- (C9);
\draw (C5) -- (C10);
\draw (C1) -- (C9);

\end{tikzpicture}
    \vspace{-1em}
    \caption{Selected Bayesian network structure. Arrows indicate conditional dependencies, with parent nodes influencing the distributions of their child nodes. Though many more characteristics and dependencies may exist, we present a simplified model for clarity. \Cref{tab:biosignatures} contains descriptions of all nodes.}
    \label{fig:bayesnet}
    \vspace{-1em}
\end{figure}

The Bayesian network in \Cref{fig:bayesnet} demonstrates how the existence of life $s_L$ directly informs the presence of various biosignatures in a sample. We employ a \textit{causal graph} structure, where parent nodes directly influence the observations of their child nodes. This structure often results in sparser connections between nodes when modeling real-world systems~\cite{koller2009pgm}. Some nodes, $o_1$ through $o_5$, represent strong biosignatures, such as the presence of polyelectrolytes or cells, while others are weaker, more ambiguous indicators of life. For example, measured pH $o_9$ (not itself a biosignature) depends on amino acid abundance $o_5$ and polyelectrolytes $o_1$, which more directly depend on whether the sample contains life. This leads to a multi-layer Bayesian network structure, with the life state $s_L$ as the root node that affects all biochemical children nodes. 

\begin{figure*}[tbp!]
    \centering
    \includegraphics[width=\linewidth]{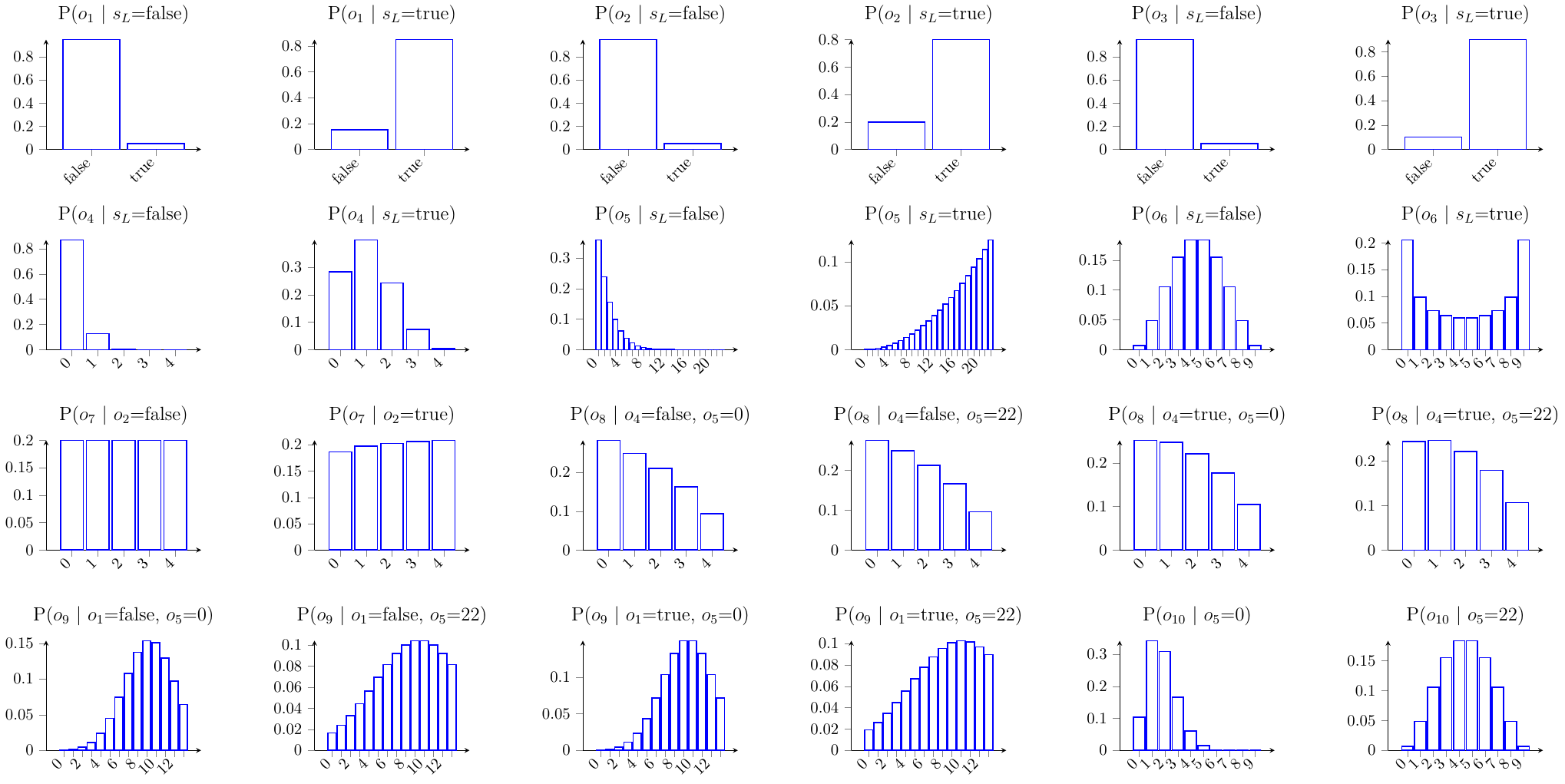}
        \caption{Full set of conditional probability distributions used in constructing the Bayesian network. Future studies may develop data-driven distributions to improve model accuracy. Some distributions exhibit sharper differences between biotic and abiotic samples, making them more discriminative biosignatures.}
    \vspace{-1em}
    \label{fig:cpds_large}
\end{figure*}

For each metric in \Cref{tab:biosignatures}, we define conditional probability distributions (CPDs) representing the likelihood of each measurement given the parent nodes. These CPDs are shown in \Cref{fig:cpds_large}. 
Each distribution is designed using insights from existing astrobiology literature, though the underlying assumptions and parameter choices can easily be adjusted to reflect various expert opinions~\cite{mathies2017feasibility,dorn2011monomer,marshall2017pathway}. Accuracy may be limited due to the lack of empirical data on biosignatures from icy moons like Enceladus. We follow the strategies for addressing the challenges of biosignature CPD design from \citet{marshall2017pathway}. As future missions collect more observational data from deep space environments, both structure and CPD parameter learning can be performed to design Bayesian networks that better represent the statistical patterns of biosignature evidence. To enable exact Bayesian inference, we discretize the CPDs using histogram binning.

\subsection{Actions}We define several actions $a$ that the spacecraft can perform after each observation $o$. These are
\begin{equation}
    \mathcal{A} = \begin{cases}
        a_{1}, \ldots, a_{6} & \text{\small Use instrument } a_i \quad \forall i \in \{1,\ldots,6\} \\
        a_7 & \text{\small Accumulate sample volume} \\
        a_8 & \text{\small Declare abiotic (terminal)}\\
        a_9 & \text{\small Declare biotic (terminal)}
    \end{cases}
\end{equation}
where instrument actions $a_1,\ldots, a_6$ can only be performed if enough sample volume is available in the funnel. For example, the Orbilander's nanopore requires far more sample volume than other instruments like the electrochemical sensor array. Sample usage values are expressed on a 0-to-1 scale, where 1 corresponds to full utilization of the sample chamber (\Cref{tab:lifeDetectionPOMDP}). Each instrument action $a_1,\ldots, a_6$ produces an observation $o$ of the current state $s$, and returns the measurements listed in \Cref{tab:instruments}. For example, HRMS collects four measurements $o_5,o_7,o_8,o_{10}$ at once. For actions with multiple observations, we perform a single belief update for each combination, capturing the joint information from all sensors.

The final actions, $a_8$ and $a_9$, are terminal declaration actions. Action $a_8$ is taken when the sample lacks evidence of life (abiotic), while action $a_9$ declares that life has been detected (biotic).

\begin{table}[htbp!]
  \centering
  \caption{Orbilander Life Detection Instruments}
  \label{tab:instruments}
  \begin{tabular}{@{} l l c @{}}
    \toprule Action &
    Instrument Name & Measurements \\ 
    \midrule
    $a_1$ & High-Resolution Mass Spectrometer (HRMS) & $o_5, o_7, o_8, o_{10}$ \\
    $a_2$ & Separation Mass Spectrometer (SMS) & $o_5, o_6$ \\
    $a_3$ & Microfluidics Device ($\mu$CE-LIF) & $o_5, o_6$ \\
    $a_4$ & Electrochemical Sensor Array (ESA) & $o_7, o_8$ \\
    $a_5$ & Microscope & $o_2, o_3$ \\
    $a_6$ & Nanopore & $o_1$ \\
    \bottomrule
  \end{tabular}
\end{table}
\subsection{Transition Function}The effects of each action $a_i$ on the sample life state $s_L$ and volume $s_V$ are captured by the transition function $T(s, a)$. Given a state-action pair $(s,a)$, $T(s, a)$ defines a probability distribution over possible next states $s'$. We define the possible state-action pair transitions on the Enceladus Orbilander's LDS as:
\begin{equation}
T(s, a) = 
\begin{cases}
    \begin{array}{l}
        s_L' \sim P(s_L) \\
        s_V' = s_V + v_{acc} \text{ where }\\ \quad  v_{acc} \sim \mathcal{N}(\mu_{acc}, \sigma^2_{acc}) \\[0.3em]
    \end{array} & \text{if } a = a_7 \\
    
    \begin{array}{l}
        s_L' = s_L \\
        s_V' = s_V - v_{use}(a) \\[0.3em]
    \end{array} & \text{if } a \in \{a_1, \ldots, a_6\} \\
    
    \begin{array}{l}
         s'  = s \text{ (terminal)} \ 
    \end{array} & \text{if } a \in \{a_8, a_9\}
\end{cases}
\end{equation}
Three types of transition may occur for a state-action pair $(s,a)$: sample accumulation $a_7$, instrument use $a_1,\ldots, a_6$, and declaration of the sample's hidden life state $a_8, a_9$. During sample accumulation, $s_V$ increases according to a Gaussian distribution centered at the plume fallout rate $v_{\text{acc}}$, or $\mathcal{N}(v_{\text{acc}}, \sigma^2)$, where $\sigma$ captures variability in the accumulation process due to environmental and operational factors. Every accumulation period has a small likelihood $P(s_L)$ of collecting a biotic sample. When using instruments, $s_L$ remains unaffected, and only sample volume $s_V$ gets reduced according to the amount consumed by instruments $a_1,\ldots, a_6$. A terminal state is reached when the sample is declared abiotic $a_8$, or biotic $a_9$.


\subsection{Reward Function}Our reward function captures several competing objectives within instrument operations. In the context of life detection, a fundamental tradeoff arises between maximizing responsiveness to biosignatures and minimizing uncertainty in the belief over the sample’s latent state $s_L$. While responsiveness allows for quick detection of potential indicators of life, extensive measurements reduce state uncertainty and ensure more confident scientific conclusions. To balance the consequences of each, we introduce a design parameter $\lambda \in [0,1]$, which scales the cost ratio between false conclusions and information gathering (dictating the agent's priorities). The full reward equation is
\begin{equation}
    R(s,a) = \begin{cases}
        0 & \text{\small Correct declaration} \\
        -\lambda & \text{\small Incorrect declaration}\\
        (1 - \lambda) \frac {s_V}{s_V^{max}}\ & \text{\small Running instrument}\\
        - \infty &\text{\small Infeasible actions}\\ 
    \end{cases}
\end{equation}
where $s_V^{max}$ represents the maximum sample volume capacity in the LDS sample preparation chamber.
 
 Infeasible actions, such as running an instrument without enough sample volume, earn a near-infinite penalty. Valid instrument uses are penalized according to how much sample volume is consumed, scaled by $(1-\lambda)$. False conclusions about a sample's biology incur a heavier penalty with larger $\lambda$ to encourage accurate declarations. A lower $\lambda$ favors less resource usage at the cost of timely reactions to potentially transient biosignatures. We perform reward tuning experiments in \cref{sec:results} to determine the optimal $\lambda$ factor.


\subsection{Solvers}To solve a POMDP, the agent maintains a belief $b$, a probability distribution over possible states, to track its uncertainty about the environment. This belief $b$ updates with each new observation $o$. Most solvers output action sequences that attempt to maximize expected cumulative reward under uncertainty. A mapping of an agent's belief to an action is known as a \textit{policy}, where $\pi(b) = a$.
We use the SARSOP algorithm~\cite{kurniawati2008sarsop} to precompute an offline policy. Offline policies are static once deployed, but they can be thoroughly tested and validated prior to deployment to provide performance guarantees. SARSOP approximates the optimal value function 
by sampling only reachable beliefs, which improves efficiency over exact methods and is well-suited to problems with large or continuous state spaces.

\subsection{Baseline}We compare our methodology to the Concept of Operations outlined in the Enceladus Orbilander's mission planning document~\cite{mackenzie2020enceladus}. We provide the full ConOps in \Cref{alg:CONOPS}. With each sampling cycle, the algorithm accumulates plume material up to a specified maximum volume $s_V^{max}$ needed to run all instruments. Once sufficient volume is collected, the algorithm sequentially activates each scientific instrument to analyze the sample. Each measurement updates the belief $b(s_L)$. Then, if the belief exceeds the life detection threshold $T_{biotic}$, the sample is declared biotic; if it falls below the abiotic threshold $T_{abiotic}$, the absence of life is declared. This threshold-driven approach imposes a rigid structure on the life detection process, potentially limiting flexibility in interpreting ambiguous or borderline results.

\begin{algorithm}[htbp!]
\caption{Enceladus Orbilander Concept of Operations for Life Detection}\label{alg:CONOPS}
\begin{algorithmic}[1]

\State\textsc{OrbilanderLifeDetection}
\State Initialize $b(s_L)$ \Comment{Belief in life}
\State Initialize $T_{biotic}$ and $T_{abiotic}$ 
\State Initialize sample volume: $s_V \gets 0$ 
\While{mission not complete}  \Comment{Repeat for entire surface mission}
\While{$T_{abiotic} < b(s_L) < T_{biotic}$}
    \State \textcolor{gray}{// Phase 1: Accumulate sample}
    \While{$s_V < s_V^{max}$}
        \State  Accumulate sample $s_V \gets s_V + v_{acc}$
    \EndWhile

    \State \textcolor{gray}{// Phase 2: Run instruments}
    \For{each instrument $a_i$ in $a_1,...a_6$}
        \State Run instrument $a_i$ and receive observation $o_i$
        \State $s_V \gets s_V - v_{use}(a_i)$
        
        \State \textcolor{gray}{// Phase 3: Update belief}
        \State Update $b(s_L) \gets P(s_L \mid o_1)$

        \State \textcolor{gray}{// Phase 4: Decision}
        \If{$b(s_L) \geq T_{biotic}$}
            \State Declare \textbf{Life Detected}
        \ElsIf{$b(s_L) \leq b_{abiotic}$}
            \State Declare \textbf{No Life Detected}
        \EndIf
    \EndFor

\EndWhile

\EndWhile

\end{algorithmic}
\end{algorithm}

\section{Results}
\label{sec:results}
We evaluated the performance of our precomputed POMDP policies against the Enceladus Orbilander ConOps. To select an effective policy, we swept over the parameter $\lambda$ to assess the tradeoff between minimizing $s_L$ uncertainty and conserving resources. We selected several policies corresponding to optimal $\lambda$ values and analyzed the resulting alpha vector plots. Finally, we compared the classification performance of the SARSOP-derived policy and the Orbilander ConOps in both nominal and off-nominal simulations. 

\begin{table}[htbp!]
  \centering
  \caption{Static POMDP parameters}
  \label{tab:lifeDetectionPOMDP}
  \begin{tabular}{@{}l c c@{}}
    \toprule
    Parameter & Variable & Value \\
    \midrule
    Number of instruments & $n$ & 6 \\
    Sample chamber capacity & $s_V^{max}$ & 1.0\\
    Sample fallout rate & $v_{acc}$ & 0.03 \\
    Sample fallout variability & $\sigma$ & $0.5v_{acc}$ \\
    \vspace{1mm}
    Instrument sample usage:& \multicolumn{2}{l}{$ s_V^{a_i}$   $\forall i\in\{1,...,6\}$} \\
    \vspace{1mm}
    \quad HRMS & $s_V^{a_1}$ & 0.01 \\
    \vspace{1mm}
    \quad SMS & $s_V^{a_2}$ & 0.06 \\    \vspace{1mm}
    \quad $\mu$CE-LIF & $s_V^{a_3}$ & 0.02 \\
    \vspace{1mm}
    \quad ESA & $s_V^{a_4}$ & 0.03 \\
    \vspace{1mm}
    \quad Microscope & $s_V^{a_5}$ & 0.01 \\
    \vspace{1mm}
    \quad Nanopore & $s_V^{a_6}$ & 0.89 \\
    \bottomrule
  \end{tabular}
\end{table}

The constants used in our experiments are listed in \Cref{tab:lifeDetectionPOMDP}. For all evaluations, samples were generated from a test Bayesian network that replaced the originally discretized conditional probability distributions in \Cref{fig:cpds_large} with continuous versions. These more closely resemble raw sensor inputs, which become discretized during post-processing.


\subsection{Reward Tuning: Selecting $\lambda$ }

\begin{figure}[bp!]
    \centering

    \includegraphics[width=\linewidth,trim={0 0 0 0},clip]{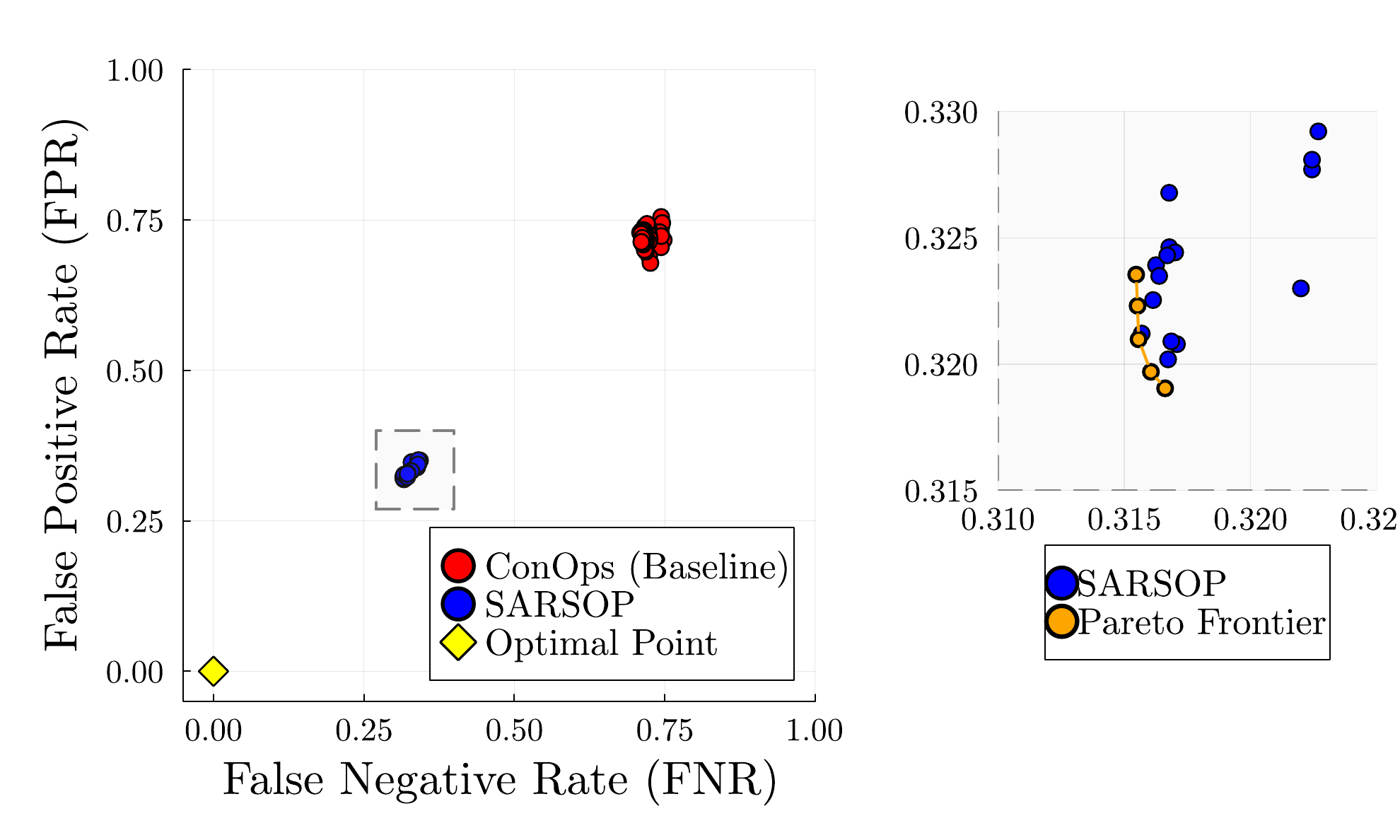}
    \vspace{-1em}
    \caption{Pareto frontier indicating optimal tradeoff between sample false positive and negative rates from a hyperparameter sweep of $\lambda \in [0.7, 1.0]$. Results for both ConOps baseline (red) and SARSOP solutions (blue) are presented, with an inset highlighting the Pareto curve for lower error regions. The inset further details the clustering of SARSOP solutions and illustrates the balance achieved along the frontier.}
    \label{fig:pareto_curve}
\end{figure}
To select an optimal policy, we compared the ConOps baseline and SARSOP-generated policies average false positive (FPR) and false negative rates (FNR) in sample classification. Our analysis focused on $\lambda \in [0.7, 1.0]$, where the policies' performance exhibited the greatest variability to hyperparameter tuning. For the ConOps, thresholds for biotic and abiotic declarations were swept over:  $T_{biotic} \in [0.9, 1.0]$ and $T_{abiotic} \in [0, 0.1]$. Each policy was evaluated over 10,000 rollouts of 200 steps. \Cref{fig:pareto_curve} displays the resulting FNR-FPR trade-offs.

All combinations of ConOps thresholds for $T_{biotic}$ and $T_{abiotic}$ led to consistently higher false negative and false positive rates in sample identification (consistently above 70\% for each metric) compared to the SARSOP-generated policies. This performance stemmed from the baseline’s inability to balance sample accumulation with opportunistic sensing. Without state-aware decisions, the ConOps approach frequently missed critical chances to detect biosignatures. In contrast, our SARSOP-generated policies made more informed, context-sensitive decisions that coordinated sensing and accumulation more effectively, with the best policies leading to a 40\% FNR and FPR reduction. This corresponded to an improvement in true positive and true negative rates from 28\% to 68\%.

The SARSOP-generated policies also spanned a broader performance range, with several configurations achieving improved trade-offs along the Pareto frontier. This flexibility arose from the role of $\lambda$ in our reward function, which governs the penalty associated with misclassification. Generally, lower $\lambda$ values encouraged risk-tolerant policies with minimal measurements. In contrast, higher $\lambda$ values resulted in more conservative policies that continued to take measurements until extremely confident in $b(s_L)$. The inset in \Cref{fig:pareto_curve}, capturing policies with $\lambda \in [0.7,0.8]$, highlights the region of interest where SARSOP solutions are tightly clustered near the Pareto frontier. The offline policies in this region achieved both better FPR and FNR than the baseline.

\begin{figure}[tbp!]
    \centering
    \includegraphics[width=0.9\linewidth]{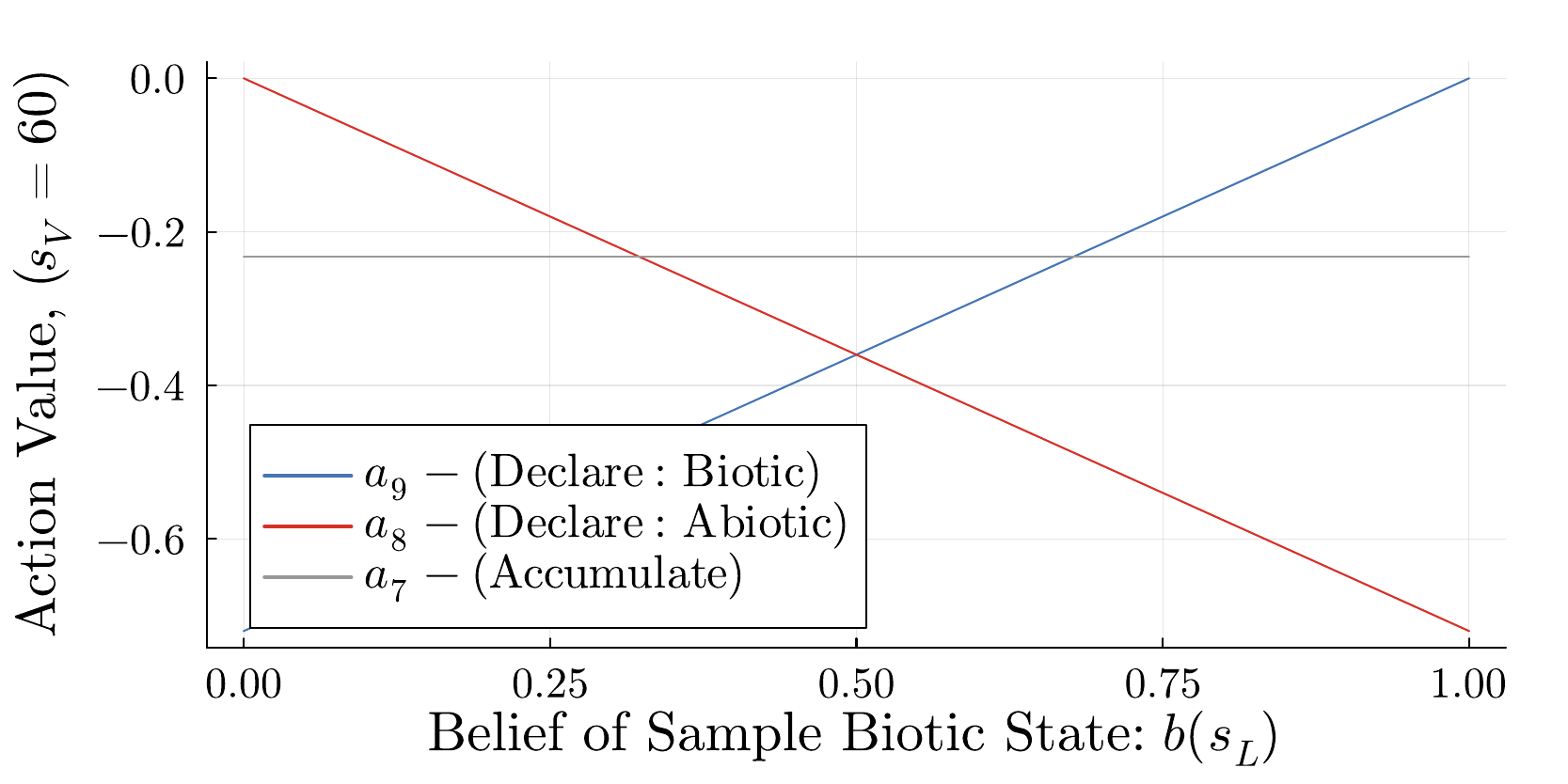}
    \caption{Example set of alpha vectors (expected utilities) for each action over belief in $s_L$ at a sample volume $s_V$ = 60\% for the $\lambda = 0.72$ generated policy. At each belief, the optimal action is the one with the highest expected utility, or action value, as determined by the SARSOP-generated policy.}
    \label{fig:alpha_sample60}
    \vspace{-0.5em}
\end{figure}

\begin{figure*}[hb]
\vspace{-1em}
    \centering
    \begin{subfigure}[b]{0.45\linewidth}
        \centering
        \includegraphics[width=\linewidth]{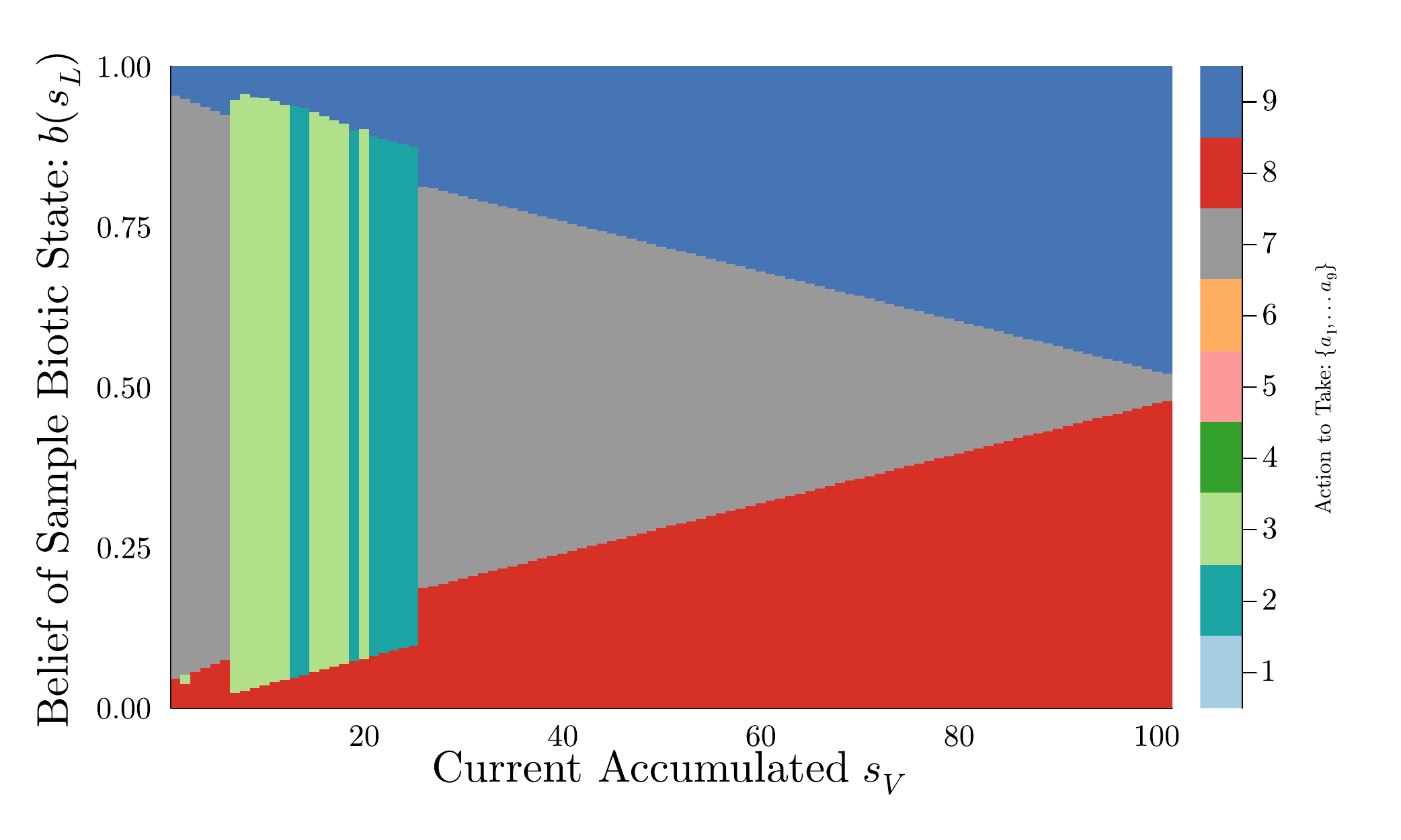}
        \label{fig:alpha1}
    \end{subfigure}
    \hfill
    \begin{subfigure}[b]{0.45\linewidth}
        \centering
        \includegraphics[width=\linewidth]{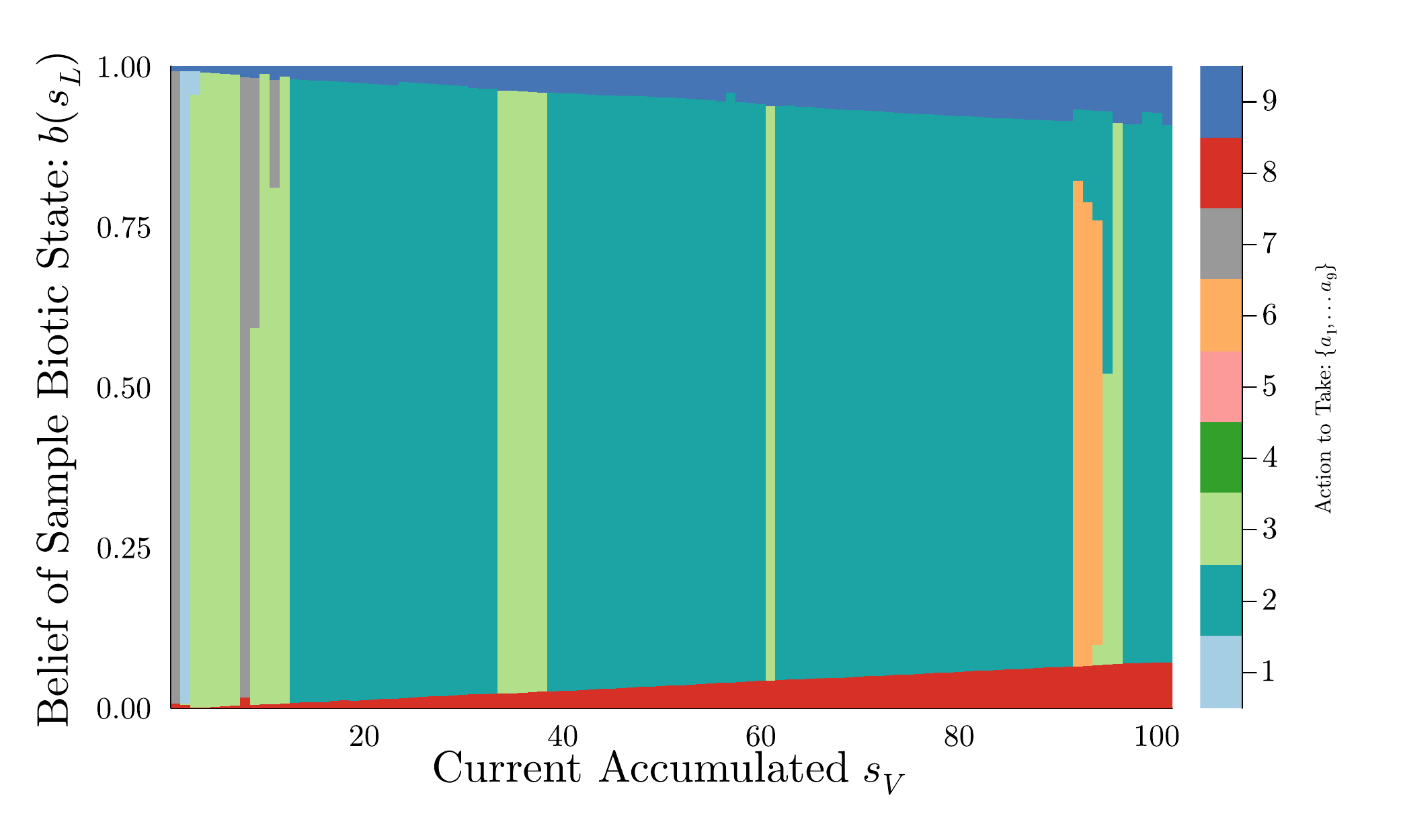}
        \label{fig:alpha2}
    \end{subfigure}
    \vspace{-2em}
    \caption{Dominating policy action at each belief state and sample volume for $\lambda = 0.72$ on the left and $\lambda = 0.935$ on the right. Legend shows actions: declare abiotic ($a_8$, red), biotic ($a_9$, blue), accumulation ($a_7$, gray), and instrument actions ($a_1, \ldots, a_6$ in other colors).}
    \label{fig:alpha_combined}
    \vspace{-1em}
\end{figure*}

Though we varied $\lambda$ to explore different reward weightings, SARSOP did not explicitly favor biotic over abiotic declarations. Consequently, increasing $\lambda$ shifted both FPR and FNR proportionally rather than revealing a clear tradeoff. This pattern was expected, given that our reward function penalized both errors equally. To induce a sharper tradeoff, the reward function would require an additional parameter to reward biotic and abiotic declarations asymmetrically.

Overall, the SARSOP-generated policies outperformed the ConOps baseline by leveraging belief-aware, adaptive decision-making. Rather than relying on fixed thresholds, these policies adapted actions based on evolving belief states. Although adjusting $\lambda$ shifted both FPR and FNR in tandem, all resulting solutions achieved lower misclassification errors than the baseline. Reward function tuning enables mission operators to design flexible policies tailored to specific operational requirements.

\subsection{Alpha Vector Policy}

To better interpret the offline policies generated with SARSOP, we examined the alpha vectors, which give the expected utility of each action from each possible state. Each alpha vector is linked with a specific action, meaning the dominating alpha vector returns the most optimal action to take under a given belief over $s_L$. Examples of dominating alpha vectors are shown in \Cref{fig:alpha_sample60}.

We visualized policies for two different reward weightings: one on the Pareto frontier at $\lambda = 0.72$ (average FNR 31.7\%, FPR 31.9\%), and another from the higher-$\lambda$ cluster at $\lambda = 0.935$, (average FNR 59.4\%, FPR 59.3\%). \Cref{fig:alpha_combined} shows the dominant alpha vector across belief in life $s_L$ and sample volume $s_V$ for each case, highlighting how the tuned reward function shapes different action sequences. This plot can be used as a lookup table of the optimal action for each potential state, as well as a validation tool to check the policy's recommendations. For $\lambda = 0.72$ in \Cref{fig:alpha_combined} (left), the agent declares samples biotic (blue) or abiotic (red) at high and low beliefs, accumulating (gray) mostly at low volumes. Instrument actions cluster at intermediate volumes, but gaps appear at larger volumes where more evidence gathering might be advantageous. This leads to seemingly premature declarations despite available sensing capacity. While this policy achieves favorable average FNR and FPR rates and lies on the Pareto frontier, its behavior is not intuitive: decisions are often made without significant use of sensors, especially when volume is still available. One possible explanation is that the policy is optimized to make decisions quickly by keeping the sample volume low, favoring early declarations once the belief crosses a moderate confidence threshold. Because the lower $\lambda$ value penalizes risk less, the policy tends to declare after using only a few instruments, favoring early commitments over thorough evidence gathering.

In contrast, $\lambda = 0.935$ in \Cref{fig:alpha_combined} (right) exhibits more conservative behavior, with instruments occupying a larger region of the action space. Although this policy has higher error rates on average, it better reflects desired operational reasoning: maximize evidence acquisition before committing to high-stakes decisions. 
By examining the resulting action maps, we can interpret how such policies would behave if deployed, which can guide further parameter tuning or reward shaping to encourage different behaviors.

\subsection{Robustness in Off Nominal Scenarios}

To assess robustness under off-nominal conditions, specifically slow or fast sample accumulation rates, we compared the two policies FNR and FPR across both scenarios in \Cref{tab:offnominal}. The results demonstrated that SARSOP-based strategies excelled at adapting to changes in sample accumulation. When accumulation was slow ($v_{acc} = 0.03$), SARSOP substantially reduced both FNR and FPR compared to the ConOps baseline; the FNR dropped from 72.4\% in the baseline to as low as 31.4\% with SARSOP. This advantage became even more pronounced under fast accumulation conditions ($v_{acc} = 0.1$), with SARSOP achieving an FNR as low as 12.9\%, a dramatic improvement over the baseline’s 42.1\%. These results show that SARSOP maintained effective detection regardless of whether sampling was unexpectedly fast or slow, underscoring its robust handling of off-nominal scenarios through adaptive decision-making.

\begin{table}[htbp]
\centering
\caption{FNR and FPR across varying $v_{acc}$.}
\label{tab:offnominal}
\begin{tabular}{@{}llcc@{}}
\toprule
\multicolumn{2}{c}{} & \multicolumn{2}{c}{\textbf{Sample Accumulation Rate}} \\
\cmidrule(lr){3-4}
\textbf{Method} & \textbf{Metric} & \textbf{Slow ($v_{acc} = 0.03$)} & \textbf{Fast ($v_{acc} = 0.1$)} \\
\midrule

\multirow{2}{*}{\makecell[l]{SARSOP\\($\lambda = 0.72$)}} 
    & FNR (\%) & 31.7 & 12.9 \\
    & FPR (\%) & 31.9 & 18.6 \\
\addlinespace

\multirow{2}{*}{\makecell[l]{SARSOP\\($\lambda = 0.935$)}} 
    & FNR (\%) & 59.4 & 17.5 \\
    & FPR (\%) & 59.3 & 17.0 \\
\addlinespace
\multirow{2}{*}{\makecell[l]{ConOps\\(Baseline)}} 
    & FNR (\%) & 72.4 & 42.1 \\
    & FPR (\%) & 68.9 & 43.3 \\
\bottomrule
\end{tabular}
\end{table}


\vspace{-1em}
\section{Conclusions}

This work presents a solution for autonomous adaptive science operations in deep space missions, using Enceladus life detection instrument operations as a case study. We proposed a novel integration of a Bayesian network to represent the observation space of biosignatures captured by science instruments, combined with a POMDP formulation to guide autonomous decision-making under uncertainty using verifiable precomputed policies, showing improvements in both false positive and negative rates of almost 40\% in comparison to the recorded baseline.

Although grounded in the Enceladus Orbilander mission, our framework can easily generalize to diverse resource-constrained, uncertain science operations. Its modular architecture can be applied to other missions by updating the Bayesian network and tuning POMDP parameters, laying the foundation for autonomous science operations that are efficient, responsive, and verifiable.
In future work, we seek to show improved performance and expansion into additional subsystems beyond the LDS suite.
We also hope to incorporate more engineering constraints into the decision-making framework, like power and memory budgets. By expanding our methods to handle degraded sensing, related subsystems, and dynamic environments, we move closer to building resilient, autonomous science platforms capable of real-time decision-making at the edge of the solar system. The complete codebase supporting this study is publicly available at \href{https://github.com/sisl/LifeDetection}{https://github.com/sisl/LifeDetection}.

\section*{Acknowledgment}

This work was supported by the NASA Ames Center Innovation Fund (CIF) under the grant ``SHERPA: Robust Precomputed Autonomy (RPA) Module," as well as the Stanford Graduate Fellowship (SGF) program.


\renewcommand*{\bibfont}{\footnotesize}
\printbibliography





\end{document}